\documentclass[runningheads]{llncs}
\usepackage{graphicx}
\usepackage{hyperref}
\usepackage{amsmath,amssymb} 
\usepackage{color}
\usepackage[width=122mm,left=12mm,paperwidth=146mm,height=193mm,top=12mm,paperheight=217mm]{geometry}

\usepackage{booktabs}

\newcommand{\ie}{{\emph{i.e.},\ }}
\newcommand{\etal}{{\emph{et al.\ }}}

\begin{document}
\pagestyle{headings}
\mainmatter
\def\ECCV18SubNumber{***}  

\newcommand{\rom}[1]{\uppercase\expandafter{\romannumeral #1\relax}}
\newcommand\blfootnote[1]{%
  \begingroup
  \renewcommand\thefootnote{}\footnote{#1}%
  \addtocounter{footnote}{-1}%
  \endgroup
}
\title{PIRM2018 Challenge on Spectral Image Super-Resolution: Dataset and Study} 

\titlerunning{PIRM2018 Challenge on Spectral Image Super-Resolution: Dataset}
\authorrunning{Mehrdad Shoeiby et al.}
\author{Mehrdad Shoeiby$^{1}$, Antonio Robles-Kelly$^{2}$, Ran Wei$^{1}$, Radu Timofte$^{3}$}
\institute{$^{1}$DATA61 - CSIRO, Black Mountain Laboratories, ACT 2601, Australia\\	
$^{2}$ Faculty of Sci., Eng. and Built Env., Deakin University, VIC 3216, Australia\\
$^{3}$Computer Vision Laboratory, D-ITET, ETH Zurich, Switzerland\\\tt{\url{mehrdad.shoeiby@data61.csiro.au}}\\\tt{\url{antonio.robles-kelly@deakin.edu.au}}\\\tt{\url{radu.timofte@vision.ee.ethz.ch}}}

\maketitle
\begin{abstract}
This paper introduces a newly collected and novel dataset (StereoMSI) for example-based single and colour-guided spectral image super-resolution. The dataset was first released and promoted during the PIRM2018 spectral image super-resolution challenge. To the best of our knowledge, the dataset is the first of its kind, comprising 350 registered colour-spectral image pairs. The dataset has been used for the two tracks of the challenge and, for each of these, we have provided a split into training, validation and testing. This arrangement is a result of the challenge structure and phases, with the first track focusing on example-based spectral image super-resolution and the second one aiming at exploiting the registered stereo colour imagery to improve the resolution of the spectral images. Each of the tracks and splits has been selected to be consistent across a number of image quality metrics. The dataset is quite general in nature and can be used for a wide variety of applications in addition to the development of spectral image super-resolution methods. 
\end{abstract}

\section{Introduction}
Imaging spectroscopy devices can capture an information-rich representation of the scene comprised by tens or hundreds of wavelength-indexed bands. In contrast with their trichromatic (colour) counterparts, these images are composed of as many channels, each of these corresponding to a particular narrow-band segment of the electromagnetic spectrum \cite{book_antonio_2012}. Thus, imaging spectroscopy has numerous applications in areas such as remote sensing \cite{msi_remote_2009,msi_remote_2010}, disease diagnosis and image-guided surgery \cite{msi_medical_2014}, food monitoring and safety \cite{msi_food_2012}, agriculture \cite{msi_agric_2015}, archaeological conservation \cite{msi_arch_2012}, astronomy \cite{msi_astronomy_2016} and face recognition \cite{msi_face_2017}.

Recent advances in imaging spectroscopy have seen the development of sensors where the spectral filters are fully integrated into the complementary metal-oxide-semiconductor (CMOS) or charge-coupled device (CCD) detectors. These are multi-spectral imaging devices which are single-shot and offer numerous advantages in terms of speed of acquisition and form-factor \cite{cdd_hyper_2013,cmos_review_2006}. However, one of the main drawbacks of these multispectral systems is the low raw spatial resolution per wavelength-indexed band in the image. Hence, super-resolving spectral images is crucial to achieving a much improved spatial resolution in these devices.


Note that, during recent years, there has been a steady improvement in the performance of example-based single image SR methods \cite{radu_methods_2017,kim_sr_2016,lai_sr_2017,Timofte_2018_CVPR_Workshops}. This is partly due to the wide availability of various benchmark datasets for development and comparison. For example, the dataset introduced by Timofte~\etal~\cite{radu_dataset_2014,radu_anchored_2013}, \cite{martin_dataset_2001,yang_srdataset_2010,zeyde_srdataset_2010,bevilacqua_srdataset_2012}, Urban100 \cite{huang_sr_dataset_2015}, and DIV2K~\cite{dataset_radu_2017} are all widely available.

Similar to RGB or grey-scale super-resolution, recently example-based techniques for spectral image super-resolution have started to appear in the literature~\cite{li_sr_spectral_2017}. However, in contrast to their RGB and grey-scale counterparts, multispectral/hyperspectral datasets suitable for the development of single image super-resolution are not as abundant or easily accessible. For example, the CNN-based method in~\cite{li_sr_spectral_2017} was developed by putting together three different hyperspectral datasets. The first of these, the CAVE~\cite{yasuma_spectral_ds_2010} consists of only 35 hyperspectral and RGB pairs gathered in a laboratory setting and controlled lighting using a camera with tunable liquid crystal filters. Similarly, the second dataset from Harvard~\cite{chakrabarti_spectral_ds_2011} contains fifty hyperspectral images captured with a time-multiplexed 31-channel camera with an integrated liquid crystal tunable filter. The third dataset is that in \cite{foster_spectral_ds_2004}, which includes 25 hyperspectral images of outdoor urban and rural scenes also captured using a tunable liquid-crystal filter. Probably the largest spectral dataset to date with more than 250 31-channel spectral images is the one introduced with the NTIRE 2018 challenge on spectral reconstruction from RGB images~\cite{Arad_2018_CVPR_Workshops}. 



Moreover, while the topic of spectral image super-resolution utilizing colour images, \ie pan-sharpening, has been extensively studied \cite{loncan_spectral-pansharp_2015,lanaras_spectral-rgb_2015,kawakami_spectral-rgb_2011} so as to develop efficient example-based super-resolution methods, stereo registered colour-spectral datasets are limited to small number of hyperspectral images. One of the very few examples is that of the datasets in \cite{dataset_msift_sabine_2011}, where the authors introduced a stereo RGB and near infrared (NIR) dataset of 477 images and propose a multispectral SIFT (MSIFT) method to register the images. However, the dataset is promoted in the context of scene recognition. In addition, the NIR images are comprised of only one wavelength-indexed band. 
Similarly, in \cite{match_nir_rgb_2018}, the authors introduce an RGB-NIR image dataset of approximately $13$ hours video with only one band dedicated to NIR images. The dataset was gathered in an urban setting by mounting the cameras on a vehicle. 

In this paper we introduce a novel dataset of colour-multispectral images which we name StereoMSI. Unlike the above two RGB-NIR datasets, the dataset was primarily developed for the PIRM2018 spectral SR challenge\footnote{Refer to \url{https://pirm2018.org/} for the spectral SR challenge and the dataset download links.} \cite{shoeiby_challenge_2018} and comprised 350 registered stereo RGB-spectral image pairs. The StereoMSI dataset is hence large enough to help develop deep learning spectral super-resolution methods. Moreover, it is, to the best of our knowledge, the first of its kind. As a result, the paper is organised as follows. We commence by introducing the dataset. We then present a number of image quality metrics over the dataset and the proposed splits for training, validation and testing. Then we present a brief review of the challenge and elaborate upon the results obtained by its participants. Finally, we discuss other potential applications of the dataset and conclude on the developments presented here.


\section{StereoMSI Dataset}
\begin{figure}
  \includegraphics[width=\linewidth]{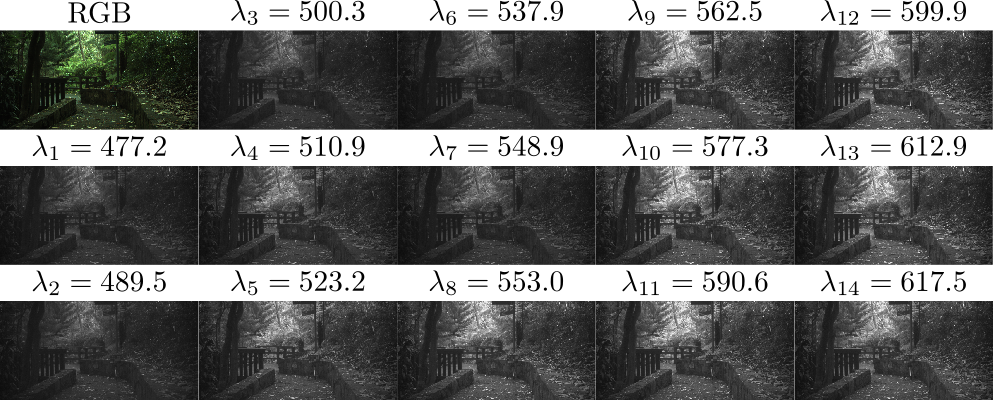}
  \caption{A sample image from the StereoMSI dataset. Here we show the RGB image and the 14 wavelengths channels of the multispectral camera indicated by $\lambda_i$, $i=\{1,2,\ldots,14\}$. All wavelengths are in $nm$ and, for the sake of better visualisation, we have gamma-corrected the 14 channels by setting $\gamma=0.75$. }
  \label{fig:dataset_01}
\end{figure}

\begin{figure}
  \includegraphics[width=\linewidth]{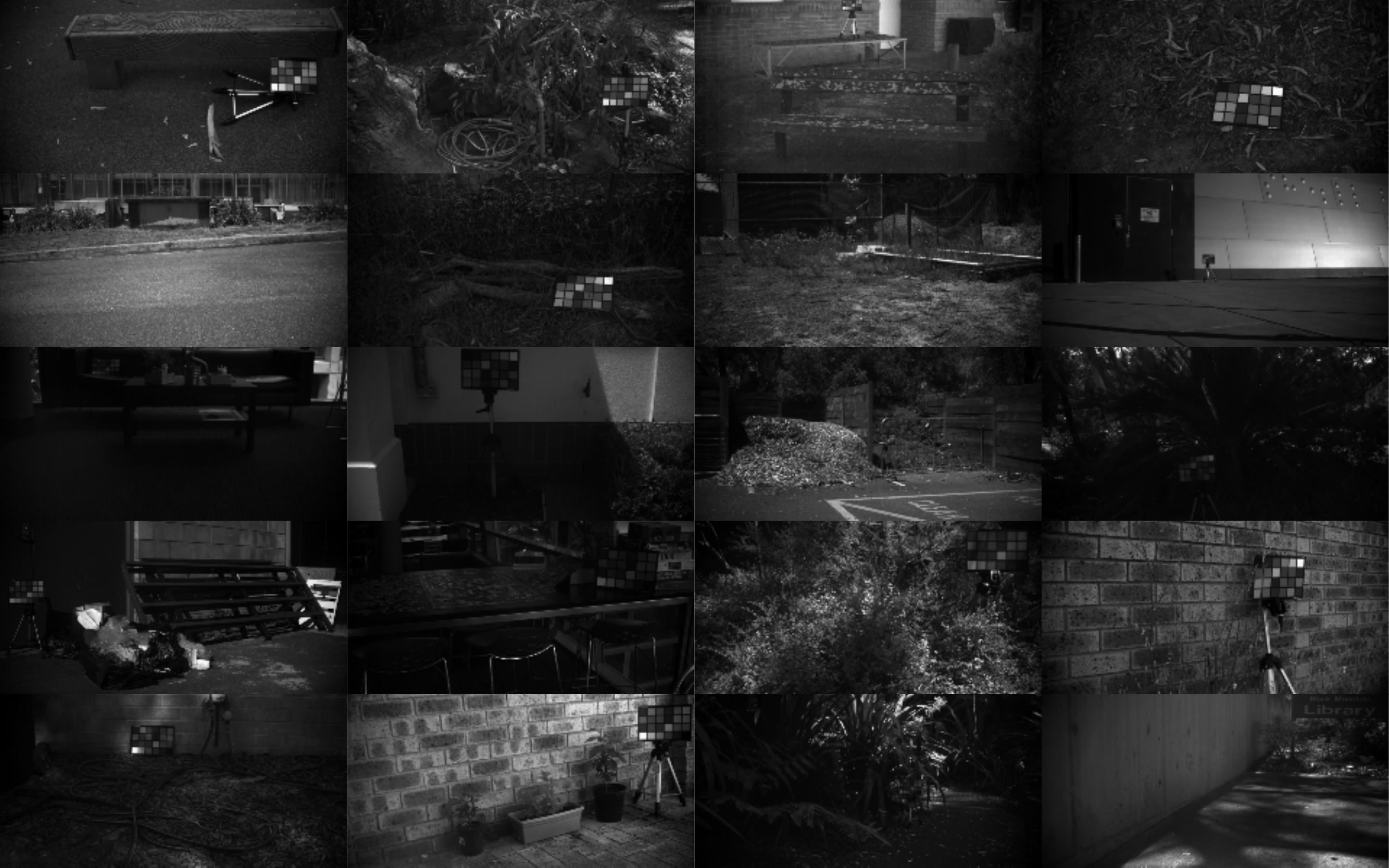}
  \caption{Validation images for the Track 1 of the PIRM2018 challenge. Each of the panels corresponds to the normalised spectral power of one of the validation images, \textit{i.e.} the norm of the spectra per-pixel normalised to unit maximum over the image.}
  \label{fig:dataset_01}
\end{figure}

\begin{figure}
  \includegraphics[width=\linewidth]{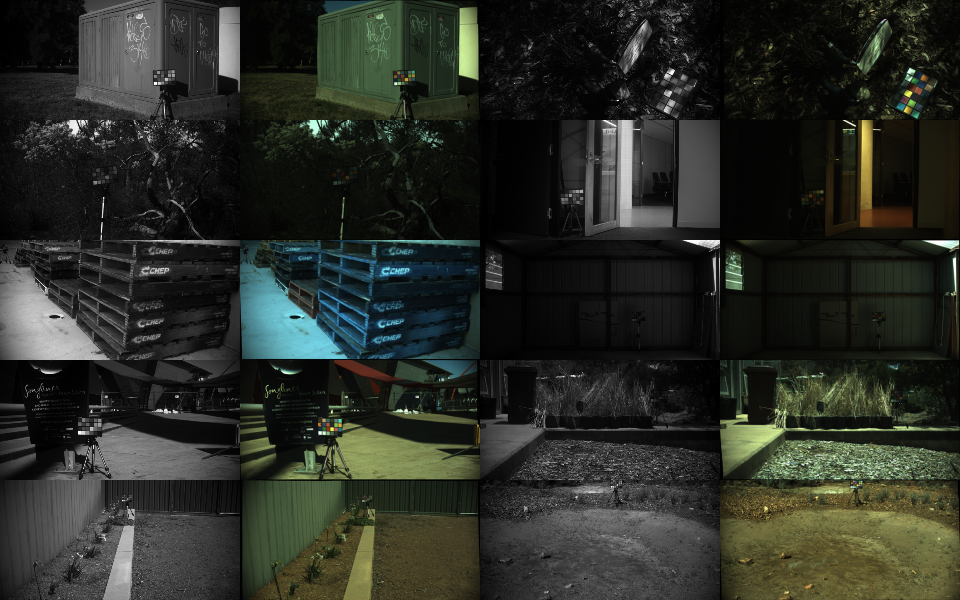}
  \caption{Validation images of Track 2 of the PIRM2018 challenge. In the left-hand and third columns we show the normalised spectral power of the spectral imagery, whereas the second and third columns show their registered RGB image pairs.}
  \label{fig:dataset_02}
\end{figure}

\begin{figure}
  \includegraphics[width=\linewidth]{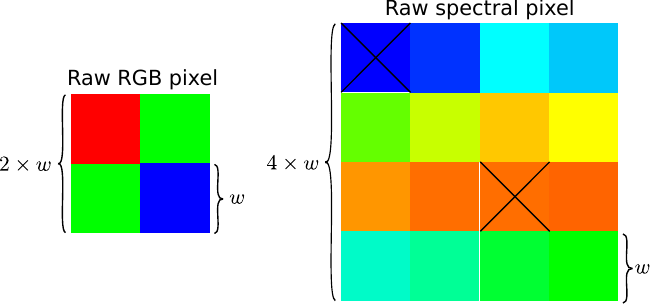}
  \caption{Illustration of raw pixels for RGB and spectral cameras. The RGB camera used to acquire the images of our dataset has $\times4$ the resolution of the spectral camera. Here we show the RGGB Bayer pattern of the colour camera and the actual wavelengths of the multispectral sensor in our MQ022HG-IM-SM4x4 camera. The two redundant filters on the array are crossed out in the panel above. The wavelengths for the all the 16 channels from the top left to the bottom right across the $4\times4$ spectral filter array are $478.2nm$, $477.2nm$, $500.3nm$, $489.5nm$, $553.0nm$, $562.5nm$, $590.5nm$, $577.3nm$, $599.9nm$, $612.9nm$, $615.9$, $617.5nm$, $510.9nm$, $523.2nm$, $548.9nm$, $537.9nm$.}
  \label{fig:raw_pixels}
\end{figure}

As mentioned above, here we propose the StereoMSI dataset. The dataset is a novel RGB-spectral stereo image dataset for benchmarking example-based single spectral image and example-based RGB-guided spectral image super-resolution methods. The dataset is developed for research purposes only.

\subsection{Diversity and Resolution} 

The 350 stereo pair images were collected from a diverse range of scenery in the city of Canberra, the capital of Australia. The nature of the images ranges from open industrial to office environments and from deserts to rainforests. In Figures \ref{fig:dataset_01} and \ref{fig:dataset_02} we display validation images for the former and latter, respectively.

It is worth noting that, during acquisition time, we paid particular attention to the exposure time and image quality as the stereo pairs were captured using different cameras. One is an RGB XiQ camera model MQ022CG-CM and the other is a XiQ multispectral camera model MQ022HG-IM-SM4x4 covering the interval $[470-620nm]$ in the visible spectral range. 


The original spectral images were processed and cropped to the resolution $480\times240$ so as to allow the stereo RGB images to be resized to a resolution 2 times larger in each axis, that is $960\times480$. This is due to the fact that, in practice, the RGB camera used, based upon a CMOS image sensor, has a $2\times2$ Bayer RGGB pattern whereas the IMEC spectral sensors have a $4\times4$ pattern delivering 16 wavelength bands. Hence, the resolution of the RGB images in each axis is twice that of the spectral images. Figure \ref{fig:raw_pixels} illustrates this resolution relationship between the two filter arrays on both cameras. When processing the images, no gamma correction was applied.


\subsection{Structure and Splits}

\begin{table}
\tiny
\begin{center}
\caption{Summarised dataset and camera properties}
\label{table:camera_properties}
\begin{tabular}{lcccccccc}
\hline
\bf{Image properties}        & \bf{StereoMSI} &\multicolumn{3}{c}{\bf{Track1}}&\multicolumn{3}{c}{\bf{Track2}}     & \bf{Testing}\\
\cmidrule(r){3-5} \cmidrule(r){6-8}
                        &     & \bf{Whole} &\bf{Training}&\bf{Validation}& \bf{Whole} &\bf{Training}&\bf{Validation}&    \\
Number of images        & 350 & 240   & 200    & 20       &   130 & 100     & 10      & 20 \\
\hline
                     &   &\multicolumn{3}{c}{\bf{Spectral}}&\multicolumn{3}{c}{\bf{RGB}} &  \\
\cmidrule(r){3-5} \cmidrule(r){6-8}
                        &     & \bf{LR}$\mathbf{\times3}$&\bf{LR}$\mathbf{\times2}$& \bf{HR}& \bf{LR}$\mathbf{\times3}$ &\bf{LR}$\mathbf{\times2}$&\bf{HR}&    \\
Image resolution &  & $(80\times160)$ &$(120\times240)$ & $(240\times480)$& $(160\times320)$& $(240\times480)$&$(480\times960)$ &\\
\hline
Spectral camera model &  \multicolumn{8}{c}{XiQ MQ022HG-IM-SM4x4}\\
RGB camera model &  \multicolumn{8}{c}{XiQ MQ022CG-CM  }\\
\hline
\end{tabular}
\end{center}
\end{table}

After collecting the StereoMSI 350 images, the two invalid wavelength-indexed bands on the IMEC sensor were removed. We then registered the images using Flownet2.0 \cite{2017_ilg_flownet} and used MATLAB's imresize\footnote{For more information on the imresize function, go to \url{https://www.mathworks.com/help/images/ref/imresize.html}} function to obtain lower resolution versions of each image by downscaling them by factors of $\times2$ and $\times3$ with nearest neighbour interpolation. 

The dataset for Track 1 (single image super-resolution) consists of 240 different spectral images.  The 240 images have been split into 200 for training, 20 for validation and 20 for testing with low resolution (HR) and high resolution (LR) on self explicatively named directories. The dataset for Track 2 (colour-guided spectral image super-resolution) consists of 120 randomly selected image stereo pairs, where one view is captured by the spectral imager and the other one by the colour camera. The images have been split into 100 pairs for training, 10 for validation and 10 for testing with HR, and LR on self explicatively named directories. All the images, for both tracks, are in band-sequential, 16 bit, ENVI standard file format. Table \ref{table:camera_properties} summarized the dataset and camera properties explained above.
\subsection{Bicubic Upsampling Metrics}

To quantitatively assess our StereoMSI dataset, and to provide a baseline for future benchmarking, we have performed image upsampling by applying a bicubic kernel.  Python's imresize function from the scikit-image\footnote{\url{https://scikit-image.org/}} toolbox was used to perform bicubic upsampling. With the upsampled images in hand, we have then computed a number of image quality metrics so as to compare the performance of current and future example-based spectral super-resolution algorithms. To this end, we up-sampled the lower-resolution images in the dataset by $\times 2$ and $\times 3$ and compared against their HR reference counterparts. 

For the sake of consistency, here we use the same metrics as those applied in the PIRM2018 spectral super-resolution challenge \cite{shoeiby_challenge_2018}. This are the mean relative absolute error (MRAE) (introduced in~\cite{Arad_2018_CVPR_Workshops}), the Spectral Information Divergence (SID), the per-band Mean Squared Error (MSE), the Average Per-Pixel Spectral Angle (APPSA), the average per-image Structural Similarity index (SSIM) and the mean per-image Peak Signal-to-Noise Ratio (PSNR). For more information on these metrics refer to the PIRM2018 spectral image super-resolution challenge report~\cite{shoeiby_challenge_2018}.

In Table~\ref{table:StereoMSI_testing}, we show the image metric results for the whole 350 images comprising the StereoMSI dataset, and the testing images. We have included the testing split in the table since the testing imagery for both tracks is the same. Table~\ref{table:tarck1}, and~\ref{table:track2} show the results for full dataset, training and validation splits of Track 1 and Track 2, respectively. 

\begin{table}
\begin{center}
\caption{Mean and standard deviation (in parenthesis) for the evaluation metrics under consideration for each of the two down sampling factors, \ie $\times2$ and $\times3$, for the whole dataset and the testing split used for both tracks of the PIRM2018 Example-based Spectral Image Super-resolution challenge.}
\label{table:StereoMSI_testing}
\begin{tabular}{|l|c|c|c|c|c|c|c|}
\hline
Dataset & Downsampling  & MRAE   & SID        & APPSA    & MSE     & PSNR & SSIM \\
split & Factor & & & & & & \\
\hline
StereoMSI    & $\times$2 & 0.28   &  0.000315  &  0.107  &  6285331  &  29.6  & 0.549\\
dataset      &           & (1.05) & (0.000329) & (0.043) & (4810190) & (3.5) & (0.062)\\
\cline{2-8}
             & $\times$3 & 0.37   & 0.000390   & 0.117    & 7779412   &  28.6  & 0.455\\
             &           & (1.62) & (0.000397) & (0.0466) & (5932423) & (3.5) & (0.069)\\
\hline
Testing      & $\times$2 & 0.18   & 0.000274   & 0.102   & 5055678   &  30.3  &  0.566\\
             &           & (0.14) & (0.000269) & (0.044) & (3481870) & (3.0) & (0.062)\\
\cline{2-8}
             & $\times$3 & 0.21    &  0.000353  &  0.110  & 6353052   &  29.3 & 0.474 \\
             &           & (0.18)  & (0.000373) & (0.047) & (4280365) & (3.1) &(0.073)\\                 
\hline
\end{tabular}
\end{center}
\end{table}

\begin{table}
\begin{center}
\caption{Mean and standard deviation (in parenthesis) for the evaluation metrics under consideration for each of the two down sampling factors, \ie $\times2$ and $\times3$, for the training and validation splits used in the Track 1 of the PIRM2018 Example-based Spectral Image Super-resolution challenge and the full set of images (the testing, training and validation splits combined).}
\label{table:tarck1}
\begin{tabular}{|l|c|c|c|c|c|c|c|}
\hline
Dataset    & Downsampling & MRAE    & SID       & APPSA    & MSE      & PSNR    & SSIM \\
split      & Factor & & & & & & \\
\hline
Full set      & $\times$2 & 0.31    &  0.000306  & 0.107   &  6054107  & 29.8   & 0.548  \\
              &           & (1.26)  & (0.000325) & (0.043) & (4672311) & (3.6)  & (0.065)\\
\cline{2-8}   & $\times$3 & 0.42    &  0.000379  & 0.117   & 7508098   & 28.8   & 0.454  \\
              &           & (1.94)  & (0.000389) & (0.047) & (5801852) & (3.6)  & (0.072)\\
\hline
Training      & $\times$2 & 0.34   & 0.000305   & 0.107   &   6228138 &  29.6  & 0.549   \\
              &           & (1.38) & (0.000331) & (0.044) & (4742463) & (3.5)  & (0.064)\\
\cline{2-8}   
              & $\times$3 &  0.45   &  0.000376  &  0.116  &  7712721  & 28.7  & 0.455  \\
              &           & (2.12)  & (0.000391) & (0.047) & (5898619) & (3.5) & (0.070)\\
\hline
Validation    & $\times$2 &  0.20   &  0.000346  &  0.115  & 5312227   & 31.0  & 0.520 \\
              &           & (0.14)  & (0.000305) & (0.038) & (4804457) & (4.5) & (0.070)\\
\cline{2-8}   
              & $\times$3 &  0.25   & 0.000430    & 0.125     & 6616913   &  30.1 & 0.421  \\
              &           & (0.21)  & (0.000378)  & (0.041)   & (5927250) & (4.6) & (0.070)\\
\hline
\end{tabular}
\end{center}
\end{table}

\begin{table}
\begin{center}
\caption{Mean and standard deviation (in parenthesis) for the evaluation metrics under consideration for each of the two down sampling factors, \ie $\times2$ and $\times3$, for the training and validation splits used in the Track 2 of the PIRM2018 Example-based Spectral Image Super-resolution challenge and the full set of images (the testing, training and validation splits combined).}
\label{table:track2}
\begin{tabular}{|l|c|c|c|c|c|c|c|}
\hline
Dataset & Downsampling    & MRAE     & SID        & APPSA    & MSE      & PSNR & SSIM     \\
split & Factor & & & & & & \\
\hline
Full dataset  & $\times$2 & 0.21  & 0.000327   & 0.108   & 6543403   & 29.3  & 0.555   \\
              &           & (0.22)& (0.000330) & (0.043) & (4873250) & (3.3) & (0.058) \\
\cline{2-8}
              & $\times$3 & 0.28  & 0.000409   & 0.118   &  8087251  & 28.3  & 0.461   \\
              &           & (0.42)& (0.000410) & 0.046   & (5937327) & (3.2) & (0.065) \\
\hline
Training      & $\times$2 &  0.20 &  0.000336  &  0.109  &  7005111  &  29.0  & 0.552  \\
              &           & (0.21)& (0.000344) & (0.044) & (5213318) & (3.4) & (0.058) \\
\cline{2-8}   & $\times$3 &  0.28 &  0.000419  &  0.119  &  8629452  &  28.0  & 0.458  \\
              &           & (0.45)& (0.000421) & (0.047) & (6352718) & (3.3) & (0.065) \\
\hline
Validation    & $\times$2 & 0.35     & 0.000337   & 0.111   & 4901777   & 29.8  & 0.558   \\
              &           & (0.32)   & (0.000284) & (0.037) & (1855890) & (1.8) & (0.045) \\
\cline{2-8}       
              & $\times$3 & 0.42     &  0.000425  &  0.121  & 61333642  & 28.8  & 0.464   \\
              &           & (0.40)   & (0.000356) & (0.041) & (2299881) & (1.7) & (0.055) \\
\hline
\end{tabular}
\end{center}
\end{table}

\section{PIRM2018 Spectral Image Super-resolution Challenge}

The PIRM2018 challenge has a twofold motivation. Firstly, the notion that, by using machine learning techniques, single image SR systems can be trained to obtain reliable multispectral super-resolved images at testing. Secondly, that by exploiting the higher resolution of the RGB images registered onto the spectral images, the performance of the algorithms can be further improved. 

Track 1 focuses on to the problem of super-resolving the spatial resolution of spectral images given training image pairs, whereby one of these is an LR and the other one is an HR image, \textit{i.e.} the ground truth reference image. The aim is hence to obtain $\times3$ spatially super-resolved spectral images making use of training imagery. Track 2, in the other hand, aims at obtaining $\times3$ spatially super-resolved spectral images making use of spectral-RGB stereo image pairs.

Each of the participating teams is expected to submit HR testing images which are to be evaluated with respect to several quantitative criteria concerning the fidelity of the reconstruction of the spectra in the super-resolved spectral images. The quantitative assessment of the fidelity of the images consists of the comparison of the restored multispectral images with their corresponding ground truth. For this, the challenge used the MRAE, the SID, the MSE, the APPSA, the SSIM and the mean PSNR. However, only MRAE and SID were used for ranking. 



\begin{table}
\begin{center}
\caption{Mean and standard deviation (in parenthesis) for the evaluation metrics under consideration for the winners (IVRL\_Prime \cite{lahoud2018multi}, and VIDAR \cite{Shi_2018}) of both tracks of the PIRM2018 Example-based Spectral Image Super-resolution challenge. For the sake of reference, we also show the results yielded by up-sampling the LR ($\times 3$) testing images using a bicubic kernel.}
\label{table:track_1_2}
\begin{tabular}{lccccccc}
\hline\noalign{\smallskip}
Team                 & Track & MRAE & SID  & APPSA & MSE    & PSNR & SSIM \\
\noalign{\smallskip}
\hline
\noalign{\smallskip}
IVRL\_Prime  & 1 & 0.07 & 0.00006 & 0.06 & 1246673 & 36.7 & 0.82 \\
VIDAR        & 1 & 0.11 & 0.00018 & 0.08 & 3414849 & 32.2 & 0.62 \\
\hline\noalign{\smallskip}
IVRL\_Prime            & 2 & 0.07 & 0.00005 & 0.05 & 852268  & 38.2 & 0.86 \\
VIDAR                  & 2 & 0.09 & 0.00011 & 0.08 & 1940939 & 34.5 & 0.75 \\
\hline
Bicubic upsampling ($\times 3$)  & & 0.21 & 0.00035 & 0.11  & 6353052   & 29.3 & 0.47\\
\hline
\end{tabular}
\end{center}
\end{table}

\begin{figure}
  \includegraphics[width=\linewidth]{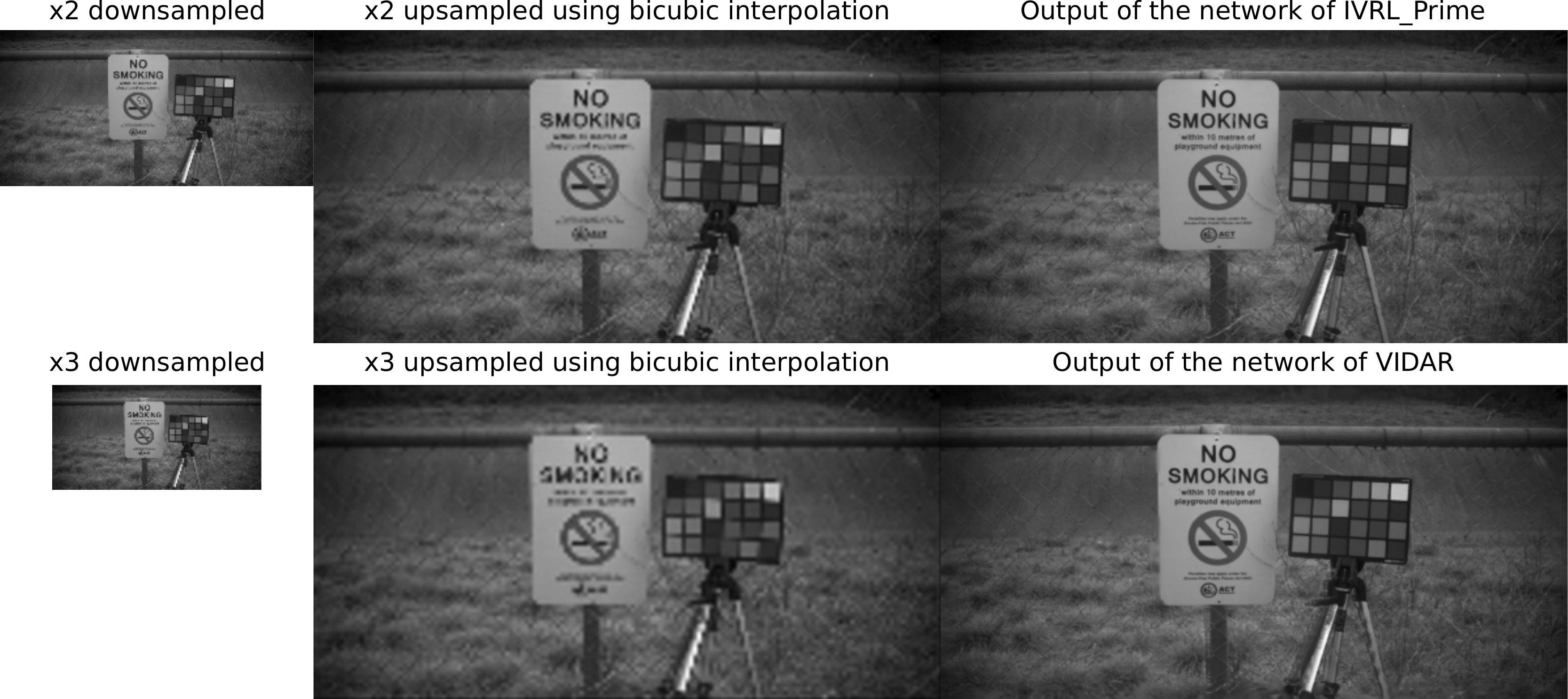}
  \caption{Performance of IVRL\_Prime, and VIDAR teams on image 124 from the Track 2 testing split, compared to bicubic upsampled LR$\times 2$ and LR$\times 3$ images. Note that, for IVRL\_Prime, inputs are LR$\times 2$ and LR$\times 3$ images, and for VIDAR the input is only the LR$\times 3$ image.  For the sake of comparison we also show an up-sampled LR image (factor $\times 3$) obtained using a bicubic kernel. All the imagery in the panels corresponds to the normalized spectral power image, and for the sake of better visualization, we have gamma-corrected the 14 channels by setting $\gamma=0.75$.
}
  \label{fig:sample_results01}
\end{figure}

In Table \ref{table:track_1_2}, we present the fidelity measurements for the testing images submitted by the challenge winners. Additionally, in Figure~\ref{fig:sample_results01} we show sample super-resolved results for the two winners of the competition. For more details regarding the challenge, the super-resolution results obtained by other participants and the networks and algorithms used at the challenge, we would like to refer the interested reader to~\cite{shoeiby_challenge_2018}.

\section{Discussion and Conclusions}

In this paper, we have introduced the StereoMSI dataset comprising of 350 stereo spectral-colour image pairs. The dataset is a novel one which is specifically structured for multispectral super-resolution benchmarking. Although it was acquired with spectral image super-resolution in mind, it is quite general in nature. Having a ColorChecker present in every image, it can also be used for a number of other learning-based applications. Moreover, it also provides lower-resolution imagery and training, validation, and testing splits for both colour-guided and example-based learning applications. We have also presented a set of quality image metrics applied to the images when up-sampled using a bicubic kernel and, in doing so, provided a baseline based upon an image resizing approach widely used in the community. We have also provided a summary of both tracks in the PIRM2018 spectral image super-resolution challenge and shown the results obtained by the respective winners.

\section*{Acknowledgments }
The PIRM2018 challenge was sponsored by CSIRO's DATA61, Deakin University, ETH Zurich, HUAWEI, and MediaTek.

\bibliographystyle{splncs}
\bibliography{references}

\end{document}